\pdfoutput=1

\documentclass[11pt]{article}
\usepackage[final]{acl}

\usepackage{times}
\usepackage{latexsym}
\usepackage{amsmath}
\usepackage{amssymb}
\usepackage[T1]{fontenc}

\usepackage[utf8]{inputenc}

\usepackage{microtype}

\usepackage{inconsolata}

\usepackage{graphicx}
\usepackage{listings} 
\usepackage{float}

\usepackage{algorithm}
\usepackage{algorithmic}
\usepackage{amsfonts}
\usepackage{multirow}
\usepackage{amsmath}

\usepackage{color,xcolor}

\title{InspireDebate: Multi-Dimensional Subjective-Objective Evaluation-Guided Reasoning and Optimization for Debating}

\author{
Fuyu Wang$^{1,2}$, Jiangtong Li$^{1,2}$\thanks{Corresponding Author.}, Kun Zhu$^{1,2}$, Changjun Jiang$^{1,2\ *}$ \\
1. Key Laboratory of Embedded System and Service Computing, \\Ministry of Education, Tongji University \\
2. School of Computer Science and Technology, Tongji University \\
\texttt{\{fywang, jiangtongli, kzhu00, cjjiang\}@tongji.edu.cn}
}

\begin{document}
\maketitle

\begin{abstract}
With the rapid advancements in large language models (LLMs), debating tasks, such as argument quality assessment and debate process simulation, have made significant progress. 
However, existing LLM-based debating systems focus on responding to specific arguments while neglecting objective assessments such as authenticity and logical validity.
Furthermore, these systems lack a structured approach to optimize across various dimensions—including evaluation metrics, chain-of-thought (CoT) reasoning, and multi-turn debate refinement—thereby limiting their effectiveness.
To address these interconnected challenges, we propose a dual-component framework: (1) \textbf{InspireScore}, a novel evaluation system that establishes a multi-dimensional assessment architecture incorporating four subjective criteria (emotional appeal, argument clarity, argument arrangement, and topic relevance) alongside two objective metrics (fact authenticity and logical validity); and (2) \textbf{InspireDebate}, an optimized debating framework employing a phased optimization approach through CoT reasoning enhancement, multi-dimensional Direct Preference Optimization (DPO), and real-time knowledge grounding via web-based Retrieval Augmented Generation (Web-RAG). 
Empirical evaluations demonstrate that \textbf{InspireScore} achieves 44\% higher correlation with expert judgments compared to existing methods, while \textbf{InspireDebate} shows significant improvements, outperforming baseline models by 57\%.
Source code is available at \url{https://github.com/fywang12/InspireDebate}.
\end{abstract}
\section{Introduction}
\begin{figure}[ht]
  \includegraphics[width=\columnwidth]{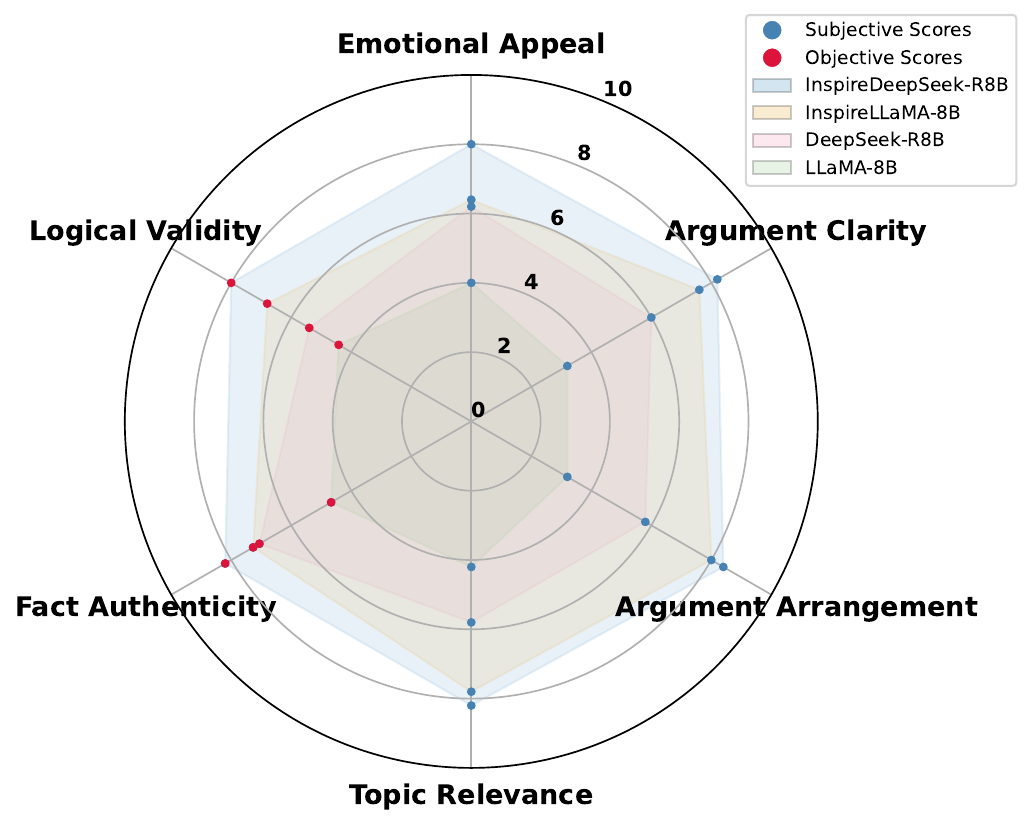}
  \vspace{-15pt}
    \caption{\textbf{Model Performance under InspireScore Evaluation.} Radar chart showing scores across four subjective (\textit{emotional appeal}, \textit{argument clarity}, \textit{argument arrangement}, \textit{topic relevance}) and two objective (\textit{logical validity}, \textit{fact authenticity}) dimensions.}
\label{fig:eval_framework}
\vspace{-15pt}
\end{figure}

In recent years, LLMs~\citep{survey_llm_Zhao_2024,survey_llmreasoning_Xu_2025,survey_llmasjudge_Gu_2025,litransformer,chi2024does} have achieved significant success in debating tasks, including argument quality assessment~\cite{Deshpande-2024—useRAGLLMforAQA} and debate process simulation~\cite{zhang_2024_Agent4debate_DBLP:journals/corr/abs-2408-04472}. 
However, systematic evaluation and optimization of structured debate processes remain underexplored.

Existing debate evaluation methods primarily assess individual arguments~\citep{Deshpande-2024—useRAGLLMforAQA} and rely heavily on subjective evaluation~\citep{Wachsmuth_AQAinLLM_bias_2024}. 
However, they lack a comprehensive assessment of the entire debate process.
Debatrix~\citep{Liang_2024ab_debatrix} advances evaluation by introducing debate-level assessment; however, it omits objective dimensionsm, such as authenticity~\citep{lehman_2019_authenticity} and logical validity~\citep{xu_2024_logical_reasoning_by_symbolic_COT_DBLP:conf/acl/Xu0P0LH24}, which are crucial for identifying misleading claims. 
Without a unified subjective-objective framework, current methods fall short in detecting hallucinations~\citep{survey_llm_hallucination_2024DBLP:journals/corr/abs-2311-05232}, an increasingly recognized challenge in LLM research.
In terms of debate frameworks, methods such as Debater~\citep{Slonim_2021_nature_IBMdebater} and Agent4Debate~\citep{zhang_2024_Agent4debate_DBLP:journals/corr/abs-2408-04472} generate fluent debates but do not provide structured representations of reasoning, which are essential for robust evaluation.

Meanwhile, MAD~\citep{liang_mad_deabte_processing_encourage_diverse_2024_acltemplate} emphasizes debate's role in enhancing reasoning, arguing that the process of debating can strengthen a model’s capacity for reasoning across downstream tasks.
For model optimization, DebateTune~\cite{debatetune_Li_2024a} enhances argument diversity but lacks evaluation-driven refinement.

To address these issues, we propose \textbf{InspireScore}, a novel evaluation system that integrates both subjective and objective dimensions to enable comprehensive debate assessment (see Figure~\ref{fig:eval_framework}).
For the subjective evaluation, we employ a structured, prompt-based assessment across key dimensions~\citep{ng-etal-2020-GAQ}, namely, emotional appeal, argument clarity, argument arrangement, and topic relevance, to ensure a nuanced, human-aligned evaluation.
For the objective evaluation, we assess both authenticity and logical validity, thereby ensuring a structured and fact-based analysis of debates.
We evaluate logical validity using first-order logic predicates and inference rules. 
In this process, LLMs convert natural language arguments into symbolic expressions and then apply logical inference to verify whether the reasoning supports the proposed conclusions.
To assess fact authenticity, we integrate LLM-based fact extraction and verification, leveraging information from external searching engine to check the authenticity of claims.
By combining these approaches, \textbf{InspireScore} offers a more reliable, comprehensive, and grounded framework for debate evaluation.

\begin{figure*}[ht]
\includegraphics[width=\textwidth]{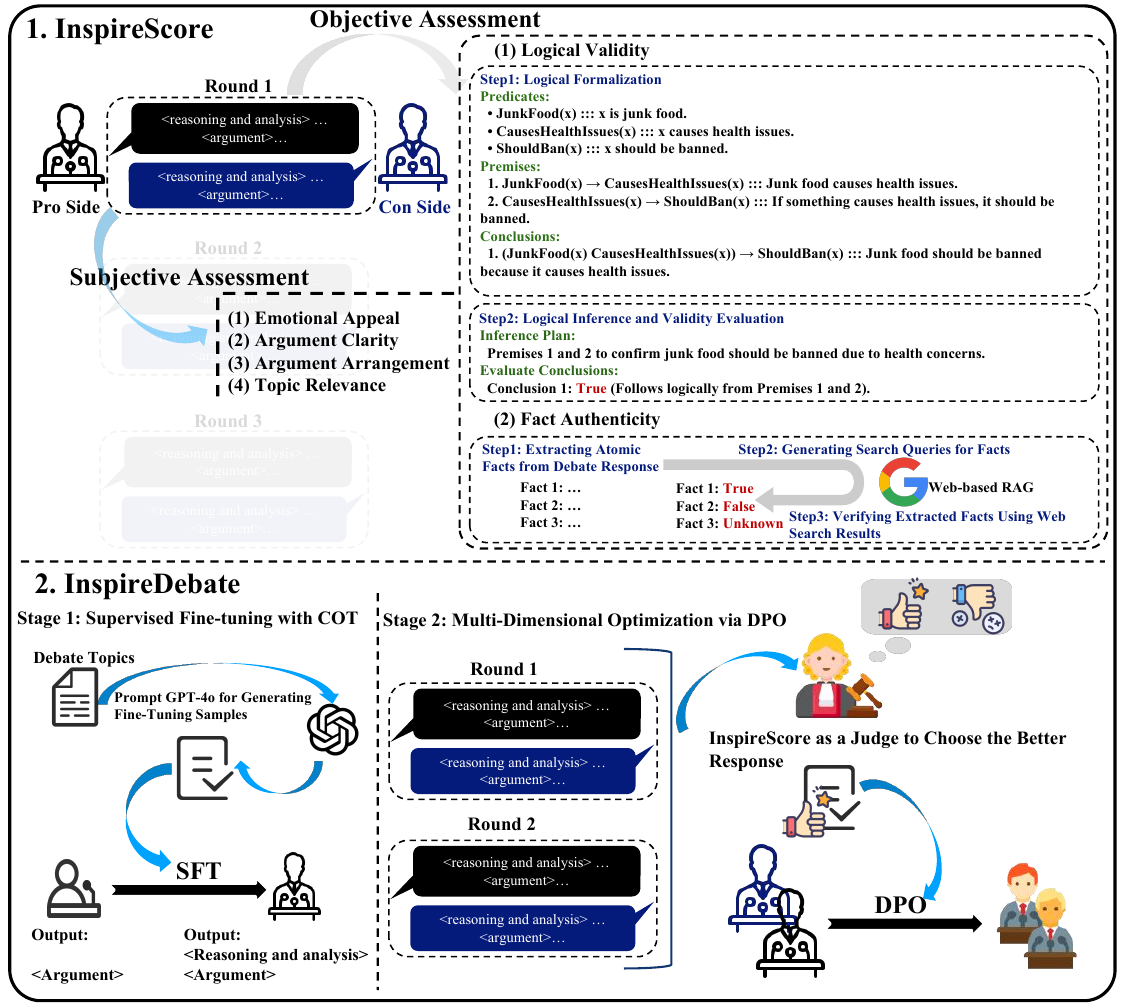}
\vspace{-14pt}
\caption{\textbf{A Unified Framework for Debating Evaluation and Optimization.} \textbf{(1) InspireScore:} Presents the evaluation structure, with subjective dimensions (\textit{emotional appeal}, \textit{argument clarity}, \textit{argument arrangement}, \textit{topic relevance}) and objective dimensions (\textit{logical validity}, \textit{fact authenticity}). \textbf{(2)InspireDebate:} Describes the two-stage optimization process: \textbf{SFT}, using GPT-4o-generated instruction data, and \textbf{DPO}, where InspireScore filters high-quality debate samples from self-debating optimization.}
\label{fig:debate_framework_version2}
\vspace{-13pt}
\end{figure*}
Building on \textbf{InspireScore}, we introduce \textbf{InspireDebate} (see Figure~\ref{fig:debate_framework_version2}), a debate optimization framework that improves LLM performance in structured debate through supervised fine-tuning (SFT)~\citep{rlhf_introsft_openai_2022} and direct preference optimization (DPO)~\citep{dpo_standford_2024}.
In the SFT stage, we construct a structured dataset that integrates the CoT process~\citep{cot_jasonwei_2022}, enabling the model to generate formalized, step-by-step reasoning outputs for more coherent debates.
In the DPO stage, we refine the model using feedback from InspireScore, aligning it with multiple evaluation dimensions rather than relying solely on static preference datasets.
To enhance real-time factual accuracy, InspireDebate incorporates Web-RAG, which enables dynamic retrieval and verification of information during debates.
These mechanisms ensure structured reasoning and adaptive optimization, making InspireDebate an effective and reliable debate optimization framework.

Experimental results show that \textbf{InspireScore} outperforms existing LLM-based evaluation systems by achieving a 44\% higher correlation with expert judgments, thereby ensuring greater consistency and comprehensiveness in structured debate assessment.
Additionally, comparative benchmarking reveals that \textbf{InspireDebate}'s iterative self-optimization improves overall debate performance by 57\%, underscoring its effectiveness.
Our key contributions are as follows:

\begin{itemize}
    \item We identify key limitations in existing debate evaluation and optimization frameworks.
    \item We propose InspireScore, improving debate evaluation from subjective-objective aspects.
    \item We introduce InspireDebate, enhancing LLM debate performance through structured reasoning and self-optimization.
    \item We validate both frameworks through experiments, demonstrating superior evaluation alignment and debate effectiveness.
\end{itemize}
\section{Related Work}
\subsection{Debate Evaluation}
Traditional debate evaluation, such as human judgment~\citep{Joshi_2023_arganalysis35k} and rule-based systems~\citep{Wang_2023_contextualForAQA,Youk_2024_highimpact}, assess arguments based on predefined criteria.
While effective, these approaches are labor-intensive, inherently subjective, require extensive annotated data, and often lack interpretability—making them unsuitable for real-time applications.

The advent of LLMs has ushered in a new paradigm for automated debate evaluation~\citep{Wachsmuth_AQAinLLM_bias_2024}.
Models like ChatGPT-4~\citep{model_gpt4_OpenAI_2024} and its variants are increasingly employed to evaluate debates, owing to their proficiency in understanding natural language, analyzing arguments, and generating coherent evaluations.
Adaptation robustness is also crucial in the large models' evaluation, and recent research has focused on the amortized evaluation for efficiency purposes in optimization~\citep{wang2025modelpredictivetasksampling}. 
Evaluating LLMs' causal reasoning measures their ability to comprehend the fundamental mechanisms that govern reality, with recent work establishing a systematic benchmark to define these capability boundaries~\cite{NEURIPS2024_af2bb2b2}.
However, current LLM-based debate evaluation systems~\citep{Deshpande-2024—useRAGLLMforAQA} fail to integrate subjective dimensions (\emph{e.g.}, emotional appeal) with objective analyses (\emph{e.g.}, logical validity), thereby limiting their capacity to deliver nuanced and comprehensive assessments~\citep{Liu_debate_eval_bias_2024b,2d-dpo_li_2024_alibaba,Mirzakhmedova_2024_areLLMAQAsAnnotators}.
Taleb~\citep{Taleb_2010_blackswan} notes that human intuition is often swayed by cognitive biases, for example, a tendency to favor concreteness over abstraction.
These observations further underscore the need for a unified debate evaluation framework that effectively integrates both subjective and objective aspects.

\subsection{Debate Framework}
Early automated debating systems, such as Project Debater~\citep{Slonim_2021_nature_IBMdebater}, have achieved significant milestones by combining argument mining, knowledge retrieval, and structured templates to generate coherent arguments.
However, these systems are limited by their reliance on curated knowledge bases and predefined rules, which restrict adaptability and scalability.
In contrast, LLM-based debating systems~\citep{zhang_2024_Agent4debate_DBLP:journals/corr/abs-2408-04472} excel in handling diverse, unstructured data and generating contextually nuanced arguments without heavy reliance on templates.
Their scalability, advanced contextual reasoning, and capacity for fine-tuning enable continuous improvement, making them better suited for complex and dynamic debating.

Despite these advantages, current LLM-based debating systems still face critical limitations.
Typically, these system lack explicit reasoning mechanisms that enhance logical coherence and depth.
Moreover, these systems typically rely on static optimization frameworks, which hinder the iterative self-optimization needed for continuous improvement and adaptability in debate performance.

\subsection{Preference Optimization}
SFT~\citep{rlhf_introsft_openai_2022} is the basis of LLM optimization, where models are fine-tuned on specific datasets to meet desired behaviors. 
However, SFT is not enough for complex logical reasoning tasks like debate. 
Reinforcement learning methods such as Proximal Policy Optimization (PPO)~\citep{ppo_openai_2017} further refine LLMs using reward signals, but PPO struggles with scalability and balancing multiple goals. 
DPO~\citep{dpo_standford_2024} provides a more efficient solution by directly optimizing preferences from reward feedback, lowering computational costs while improving task alignment. 
Recent work on multi-dimensional optimization~\citep{Liang_2024ab_debatrix} considers trade-offs across different metrics. 
Although promising, these methods are still new and often lack ways to handle biases or support continuous self-improvement needed for debate. 
MAD~\citep{liang_mad_deabte_processing_encourage_diverse_2024_acltemplate} boosts logical reasoning by allowing models to debate over several rounds, however, it lacks a clear guidance towards on objective feedback.

Although significant progress has been made in debating, current approaches lack a unified framework that integrates logical reasoning (\emph{e.g.}, CoT) with self-optimization capabilities. 
These gaps motivate the development of our InspireScore and InspireDebate frameworks, which aim to advance both debate evaluation and LLM optimization.
\section{Evaluation Framework}
Debate evaluation is an inherently complex endeavor, as it necessitates a delicate balance between subjective perceptions and objective criteria across multiple dimensions. 
Existing evaluation frameworks predominantly concentrate on subjective aspects, such as emotional appeal, clarity, and relevance, often neglecting the critical role of objective measures. 
Notably, objective criteria, including logical validity and factual authenticity, are indispensable for constructing a comprehensive evaluative system. 
These two paradigms emphasize distinct attributes: subjective evaluation privileges rhetorical artistry, whereas objective assessment underscores logical coherence and authenticity. 
This divergence can engender conflicts, exemplified by scenarios in which a debate exhibits robust logical structure yet lacks emotional resonance, or vice versa. 
To facilitate a fair, holistic assessment, we propose InspireScore, a unified framework that integrates both subjective and objective aspects.

\subsection{Subjective Evaluation}
\begin{table*}

\centering
\resizebox{\textwidth}{!}{%
\begin{tabular}{c|c|l}
\hline
Aspect &
  Dimension &
  Description \\ \hline
\multirow{7}{*}{Subjective} &
  Emotional Appeal~(EA) &
  \begin{tabular}[c]{@{}l@{}}Evaluates whether the argument evokes a sense of approval or emotional \\ resonance in the audience or judges, enhancing its persuasiveness.\end{tabular} \\ \cline{2-3} 
 &
  Argument Clarity~(AC) &
  \begin{tabular}[c]{@{}l@{}}Assesses whether the argument is expressed in a way that is clear, concise, \\ and easy for the audience or judges to understand.\end{tabular} \\ \cline{2-3} 
 &
  Argument Arrangement~(AA) &
  \begin{tabular}[c]{@{}l@{}}Evaluates whether the order and structure of the argument contribute to the \\ presentation of the viewpoints.\end{tabular} \\ \cline{2-3} 
 &
  Topic Relevance~(TR) &
  \begin{tabular}[c]{@{}l@{}}Determines whether the argument directly aligns with and addresses the debate \\ topic, ensuring its pertinence.\end{tabular} \\ \hline
\multirow{2}{*}{Objective} &
  Fact Authenticity~(FA) &
  \begin{tabular}[c]{@{}l@{}}Evaluates the proportion of independent facts in a debate that are verified as true.\end{tabular} \\ \cline{2-3} 
 &
  Logical Validity~(LV) &
  Assesses whether the reasoning in a debate logically supports the argument. \\ \hline
\end{tabular}%
}
  \caption{Evaluation Dimensions of InspireScore}
  \label{tab:table1}
  \vspace{-14pt}
\end{table*}

Subjective evaluation focuses on qualitative criteria that assess the presentation, construction, and reception of debate arguments.
Existing studies~\citep{Lauscher_2020_theory_basedonAQA_DBLP:conf/coling/LauscherNNT20, Wachsmuth_2017_AQAinNLP_DBLP:conf/eacl/WachsmuthSHPBHN17, ng-etal-2020-GAQ} propose taxonomies for evaluating argument quality, emphasizing logical cogency, rhetorical effectiveness, and dialectical reasonableness.
Debatrix~\citep{Liang_2024ab_debatrix} evaluates debates based on argument, source, and language dimensions.
Building on these works, our system refines subjective evaluation by incorporating four dimensions: emotional appeal, argument clarity, argument arrangement, and topic relevance.
Descriptions of these dimensions are provided in Table~\ref{tab:table1}.

To enable effective subjective evaluation with InspireScore, we design prompts to assess both sides of the debate across four dimensions.
The prompts are provided in Appendix~\ref{apdx:appendixA1}.
The score for each evaluation dimension is computed as follows:
\begin{equation}
S_{D} = \frac{1}{m}\sum_{i=1}^{m} s_{i,D},
\end{equation}
where $D \in \{EA, AC, AR, TR\}$ denotes the evaluation dimension, corresponding to emotional appeal, argument clarity, argument arrangement, and topic relevance, respectively. 
Here, $s_{i,D} \in [0, 1]$ and $m$ are the score of the $i$-th debate round for dimension $D$, and the number of rounds in debate.

\subsection{Objective Evaluation}
In light of the practical requirements of debate scenarios and the strengths of LLMs, we have established two key evaluation metrics within our objective evaluation system: logical validity and fact authenticity.
Structuring our system around these dimensions directly addresses the fundamental requirements of debates, ensuring both high-quality evidence and sound reasoning.
Simultaneously, we leverage the capabilities of LLMs in fact verification and logical analysis~\citep{survey_llmreasoning_Xu_2025}, thereby enhancing the objectivity, depth, and efficiency of the evaluation process.
This objective framework complements subjective evaluation, establishing a scientific and holistic debate evaluation system.

For fact authenticity assessment in debates, we optimize the solution provided by SAFE~\citep{wei_2024_SAFEforAuthenticity_DBLP:journals/corr/abs-2403-18802}.
We decompose debate responses into independent factual claims and leverage an LLM, in conjunction with external evidence, to evaluate their authenticity.
Specifically, by using information retrieved through web search engine, we verify the authenticity of each factual claim.
The prompts for fact extraction and verification are provided in Appendix~\ref{apdx:appendixA2}.
The final evaluation metric is defined as the proportion of independent facts judged to be true among all independent facts:
\begin{equation}
S_{FA} = \frac{\sum_{i=1}^{m}f_{i}(y)}{\sum_{i=1}^{m}NF_{i}},
\end{equation}
where $f_{i}(y)$, $NF_{i}$, and $m$ are the number of supported facts, the number of independent facts in $i$-th debate round and the number of rounds in debate.

To assess logical validity, we adopt a two-step approach that ensures a structured and rigorous evaluation. 
First, we convert the debate response into first-order logic (FOL) symbolic representations, thereby formalizing its underlying reasoning structure.
Second, we apply logical inference rules to these symbolic expressions, systematically verifying whether each step logically follows and leads to the final argument.
This process enables a precise assessment of logical coherence, ultimately yielding a logical validity score that reflects the consistency and soundness of the debater’s reasoning. Inspired by SymbCOT~\citep{xu_2024_logical_reasoning_by_symbolic_COT_DBLP:conf/acl/Xu0P0LH24}, our structured pipeline enhances debate evaluation by integrating symbolic logic with automated reasoning.
Detailed prompts and illustrative examples are in Appendix~\ref{apdx:appendixA3}.
The evaluation metric for logical validity is defined as the proportion of final argument expressions that can be correctly derived through formal logical inference:
\vspace{-6pt}
\begin{equation}
S_{LV} = \frac{\sum_{i=1}^{m}\sum_{j=1}^{N_i} v\left(\text{FOL}_{i}^{j}\right)}{\sum_{i=1}^{m} N_i},
\vspace{-6pt}
\end{equation}
where $N_i$, $\text{FOL}_{i}^{j}$, and $v(\cdot)$ are the number of argument expressions, the $j$-th argument expression in the $i$-th debate round, and a boolean function returning $1$ if correctly derived, and $0$ otherwise.
\section{InspireDebate Framework}
The InspireDebate framework enhances LLM-based debating capabilities by integrating supervised fine-tuning with CoT reasoning and multi-dimensional optimization via Direct Preference Optimization. This two-stage process equips the model to engage in structured debate and refinement based on comprehensive evaluation feedback. 

\subsection{SFT with CoT Integration}
Recent advancements in system-2 thinking~\citep{evans2003two_system_2_thinking,shleifer2012psychologists_empirical_evidence_fastandslowthinking}, which emphasize deliberate, analytical reasoning, have explored integrating step-by-step reasoning paradigms, such as CoT~\citep{cot_jasonwei_2022} and Rephrase and Respond~\citep{deng2023rephrase_rar_rephraseandresponse}, to enhance logical inference capabilities.
However, significant challenges remain when using open-source base models for adversarial content generation, especially in debate scenarios. These challenges are primarily due to refusal behaviors, illustrated in Appendix~\ref{apdex:B_instruction_refusal}, which result from safety fine-tuning~\citep{debatetune_Li_2024a}.

To address refusal behaviors in open-source models and leverage insights from CoT-based reasoning, we design a debate template that integrates both the \textbf{reasoning process} and the \textbf{argument output}.
This template ensures that responses include not only direct answers but also a structured breakdown of the argumentation process.
Using OpenAI GPT-4o~\citep{model_OpenAI4o_2024}, we construct a fine-tuning dataset, denoted as $D = \{(x_i, y_i)\}_{i=1}^{N}$, comprising pairs of input debate prompts $x_i$ and their corresponding structured responses $y_i$.
Training on this dataset enables the model to generate responses that not only deliver conclusions but also articulate a clear, structured reasoning process, systematically decomposing arguments rather than relying on heuristic or unstructured reasoning.
The fine-tuning objective is to maximize the conditional probability of generating the structured response $y$ given the input $x$, which can be formulated as:
\vspace{-5pt}
\begin{equation}
L_{\text{SFT}} = \mathbb{E}_{(x, y) \sim D} \left[ -\log P_{\theta}(y | x) \right],
\vspace{-5pt}
\end{equation}
where $P_{\theta}(y | x)$ denotes the conditional probability of model generating $y$ given the input $x$.

By optimizing this loss function, the model updates its parameters $\theta$ via gradient descent, thereby increasing the likelihood of generating the desired structured output.
Ultimately, this SFT process enhances the reasoning capabilities of LLMs, enabling them to produce debate responses that integrate both the \textbf{reasoning process} and the \textbf{argument output}. 
This integration is particularly crucial for effective debate analysis and assessment. 

\subsection{Multi-Dimensional Optimization via DPO}
The debate process is collaboratively driven by two LLMs, each assuming the role of either the affirmative or negative side to argue a specific topic~\citep{liang_mad_deabte_processing_encourage_diverse_2024_acltemplate}.
Before presenting their arguments, each side engages in a structured analytical reasoning process to thoroughly consider their stance.
Each debate consists of multiple rounds, the number of which can be manually configured based on experimental needs. To further enhance the explanation of the debate process, we provide an example of a debate process in Appendix~\ref{apdx:debate_example}.

To align the model with real-world debate evaluation, we introduce DPO~\citep{dpo_standford_2024} for model optimization~\citep{zheng2024llamafactory}.
Unlike conventional reinforcement learning with human feedback~(RLHF)~\citep{rlhf_introsft_openai_2022}, which optimize the LLMs based on implicit user preferences, we leverage InspireScore to fine-tune the model with explicit, multi-dimensional feedback.

The optimization framework is built on the SFT stage, ensuring that the model generates both the \textbf{reasoning process} and the \textbf{argument output}.
Following this, the model is fine-tuned using DPO, during which debate responses are iteratively refined based on evaluation scores provided by InspireScore.
Unlike conventional methods that rely on an explicitly learned reward function, InspireScore directly compares debate responses across multiple evaluation dimensions, both subjective and objective, thereby enabling a more structured and interpretable optimization process.

Given a DPO dataset containing debate samples evaluated by InspireScore, we construct preference pairs $(y_{\text{w}}, y_{\text{l}})$, where $y_{\text{w}}$ (the winning ones) achieves a higher InspireScore than $y_{\text{l}}$ (the losing ones).
The optimization objective is formulated as:
\begin{equation}
\begin{split}
&\mathcal{L}_{DPO}(\pi_{\theta}; \pi_{\text{SFT}}) = -\mathbb{E}_{(x, y_w, y_l) \sim \mathcal{D}}\\
&\left[ \log \! \sigma \! \left(\! \log \! \frac{\pi_{\theta}(y_{w} | x)}{\pi_{\text{SFT}}(y_{w} | x)} \!-\! \log \! \frac{\pi_{\theta}(y_{l} | x)}{\pi_{\text{SFT}}(y_{l} | x)} \!\right) \right],
\end{split}
\end{equation}
where $\pi_{\theta}(y|x)$ denotes the fine-tuned policy model’s probability of generating response $y$, $\pi_{\text{SFT}}(y|x)$ represents the reference model’s probability after SFT, and $\sigma(\cdot)$ is the sigmoid function.

\subsection{Real-Time Factuality with Web-RAG}
To further enhance the factual reliability of LLMs, InspireDebate integrates Web-RAG, enabling the model to dynamically retrieve and incorporate external evidence during debating. This mechanism ensures that the model accesses up-to-date factual information, reducing reliance on static knowledge and thereby improving the credibility of response.

The process involves two key steps. 
First, the model extracts relevant keywords based on the debate topic, and, when applicable, the opposing side’s arguments, which are then used to query web sources for pertinent information. Second, the retrieved information is integrated into the analysis and generation of the final argument, ensuring that the debate response remains factually consistent.
We provide an example in Appendix~\ref{appendix_rag_prompt} to 
demonstrate keyword extraction and argument generation.
\section{Experiments and Results}
\subsection{Experiments Setting} 
\textbf{Compared Method.} We conduct experiments with four open-source LLMs and two proprietary LLMs. 
For the open-source LLMs, we optimize them using InspireDebate framework, resulting in their Inspire-versions.
The open-source models include, 1) LLaMA-3.1-8B-Instruct~\citep{llama3modelcard}~(\textbf{LLaMA-8B}); 2) DeepSeek-R1-Distill-LLaMA-8B~\citep{model_DeepSeekr1_2025}~(\textbf{DeepSeek-R8B}); 3) Phi-3.5-mini-Instruct (3.6B)~\citep{abdin2024phi_model_phi35}~(\textbf{Phi-3.6B}), and 4) Qwen-2.5-1.5B-Instruct~\citep{yang2024qwen2_model_qwen25-15B}~(\textbf{Qwen-1.5B}).
The open-source LLMs are trained in two stage on 2 NVIDIA-V100 (32G) GPUs for about 2 to 3 hours.
The proprietary models include GPT-4o-mini~\citep{model_OpenAI4o_2024} and o1-mini~\citep{jaech2024openai_model_o1}. 

\noindent\textbf{Dataset.} Our experimental framework leverages a corpus of 710 debate topics drawn from a predefined library~\citep{debatetune_Li_2024a}, which are strategically allocated across different stages of our methodology.
Specifically, 100 topics are used to generate instructional examples with OpenAI GPT-4o (2024-08-06) for SFT stage.
An additional 510 topics are employed in debate simulations to construct a preference dataset for DPO stage, thereby refining model performance based on InspireScore.
The remaining 100 topics are reserved for evaluating the effectiveness of both InspireScore and InspireDebate.
During both the SFT and DPO training stages, models are trained for three epochs with a learning rate of $1 \times 10^{-5}$. 
We apply LoRA for effective fine-tuning with reduced computational overhead.
Besides, during the experimental phase, we employ the Serper API for web search.

\begin{table}
  \centering
  \resizebox{1\columnwidth}{!}{%
  \begin{tabular}{lccc}
    \hline
    \textbf{Evaluation Framework} & \textbf{Pearson} & \textbf{Spearman} & \textbf{Kendall} \\
    \hline
DAGN with CL             & 0.045   & 0.016    & 0.013   \\
GPT-4o (2024-08-06)      & 0.112   & 0.143    & 0.137   \\
GPT-4o-mini (2024-07-18) & 0.118   & 0.134    & 0.129   \\
SPARK                    & 0.019   & 0.037    & 0.032   \\
Debatrix (gpt-3.5-turbo) & 0.394   & 0.412    & 0.375   \\
\textbf{InspireScore (ours)}     & \textbf{0.643}   & \textbf{0.581}    & \textbf{0.479} \\
    \hline
  \end{tabular}
  }
  \vspace{-5pt}
  \caption{Comparison of Evaluation Frameworks against Human Judgements.}
  \label{tab:camera_ready_main_result_evaluation_framework}
\end{table}
\begin{table}
    \centering
    \footnotesize 
    \setlength{\tabcolsep}{14pt} 
    \begin{tabular}{lc}
    \hline
    \textbf{Model} & \textbf{RMSE} \\
    \hline
    ChatGPT       & 49.99 \\
    GPT-4    & 44.84 \\
    Debatrix & 42.21 \\
    \textbf{InspireScore}    & \textbf{35.36} \\
    \hline
    \end{tabular}
  \caption{Model Performance on Human Debate Winner Prediction.}
  \label{tab:camera_ready_main_result_inspire_humandebate}
  \vspace{-11pt}
\end{table}
\begin{table*}[ht]
\centering
\setlength\tabcolsep{16pt}
\resizebox{1\textwidth}{!}{%
\begin{tabular}{lcccccccccc}
\hline \hline
\multirow{2}{*}{\textbf{Model Setting}}     & \multicolumn{5}{c}{\textbf{Subjective}}                                                                                                                                      & \multicolumn{1}{c}{\textbf{}} & \multicolumn{3}{c}{\textbf{Objective}}                                                                   & \multirow{2}{*}{\textbf{InsightScore}} \\ \cline{2-6} \cline{8-10}
 & \multicolumn{1}{c}{\textbf{EA}} & \multicolumn{1}{c}{\textbf{AC}} & \multicolumn{1}{c}{\textbf{AA}} & \multicolumn{1}{c}{\textbf{TR}} & \multicolumn{1}{c}{\textbf{Average}} & \multicolumn{1}{c}{\textbf{}} & \multicolumn{1}{c}{\textbf{FA}} & \multicolumn{1}{c}{\textbf{LV}} & \multicolumn{1}{c}{\textbf{Average}} &                                        \\ \hline
Qwen-1.5B              & 0.421       & 0.420       & 0.360       & 0.283       & 0.371            &  & 0.621       & 0.388       & 0.505            & 0.416                                    \\
Inspire-Qwen-1.5B      & 0.781       & 0.780       & 0.640       & 0.705       & 0.727            &  & 0.786       & 0.720       & 0.753            & 0.735                                    \\ \hline
LLaMA-8B               & 0.403       & 0.320       & 0.320       & 0.421       & 0.366            &  & 0.467       & 0.428       & 0.448            & 0.393                                    \\
Inspire-LLaMA-8B       & 0.641       & 0.760       & 0.802       & 0.781       & 0.746            &  & 0.727       & 0.680       & 0.704            & 0.732                                   \\ \hline
Phi-3.6B               & 0.522       & 0.503       & 0.482       & 0.480       & 0.497            &  & 0.621       & 0.454       & 0.538            & 0.510                                   \\
Inspire-Phi-3.6B       & \textbf{0.803}       & 0.806       & 0.840       & \textbf{0.860}       & \textbf{0.827}            &  & 0.813       & 0.494       & 0.654            & 0.769                                   \\ \hline
DeepSeek-R8B           & 0.626       & 0.603       & 0.580       & 0.580       & 0.597            &  & 0.706       & 0.640       & 0.673            & 0.623                                    \\ 
Inspire-DeepSeek-R8B   & 0.801       & \textbf{0.823}       & \textbf{0.841}       & 0.822       & 0.822            &  & \textbf{0.820}       & \textbf{0.800}       & \textbf{0.810}            & \textbf{0.818}                                    \\ \hline
\hline
o1-mini                & 0.844       & 0.783       & 0.780       & 0.806       & 0.803            &  & 0.805       & 0.760       & 0.783            & 0.796                                    \\
GPT-4o-mini            & 0.827       & 0.880       & 0.840       & 0.860       & 0.852            &  & 0.831       & 0.801       & 0.816            & 0.840                                    \\ \hline \hline
\end{tabular}%
}
\caption{Comparison of Optimized LLMs Using InspireScore. Best open-sourced LLMs are highlighted in bold.}
\label{tab:main_result_debate_framework_version3}
\end{table*}

\begin{table*}[ht]
\centering
\setlength\tabcolsep{16pt}
\resizebox{1\textwidth}{!}{%
\begin{tabular}{lcccccccccc}
\hline \hline
\multirow{2}{*}{\textbf{Model Setting}}     & \multicolumn{5}{c}{\textbf{Subjective}}                                                                                                                                      & \multicolumn{1}{c}{\textbf{}} & \multicolumn{3}{c}{\textbf{Objective}}                                                                   & \multirow{2}{*}{\textbf{InsightScore}} \\ \cline{2-6} \cline{8-10}
 & \multicolumn{1}{c}{\textbf{EA}} & \multicolumn{1}{c}{\textbf{AC}} & \multicolumn{1}{c}{\textbf{AA}} & \multicolumn{1}{c}{\textbf{TR}} & \multicolumn{1}{c}{\textbf{Average}} & \multicolumn{1}{c}{\textbf{}} & \multicolumn{1}{c}{\textbf{FA}} & \multicolumn{1}{c}{\textbf{LV}} & \multicolumn{1}{c}{\textbf{Average}} &                                        \\ \hline
Qwen-1.5B              & 0.489       & 0.466       & 0.254       & 0.310       & 0.380            &  & 0.617       & 0.449       & 0.533            & 0.431                                    \\
Inspire-Qwen-1.5B      & 0.814       & 0.742       & 0.695       & 0.770       & 0.755            &  & 0.782       & 0.611       & 0.697            & 0.736                                    \\ \hline
LLaMA-8B               & 0.323       & 0.275       & 0.432       & 0.472       & 0.376            &  & 0.446       & 0.350       & 0.398            & 0.383                                    \\
Inspire-LLaMA-8B       & 0.683       & 0.860       & 0.693       & \textbf{0.859}       & 0.774            &  & 0.730       & 0.594       & 0.662            & 0.737                                   \\ \hline
Phi-3.6B               & 0.483       & 0.567       & 0.545       & 0.570       & 0.541            &  & 0.644       & 0.494       & 0.569            & 0.551                                   \\
Inspire-Phi-3.6B       & 0.842       & 0.846       & \textbf{0.846}       & 0.825       & 0.840            &  & 0.696       & 0.512       & 0.604            & 0.761                                   \\ \hline
DeepSeek-R8B           & 0.511       & 0.487       & 0.603       & 0.521       & 0.531            &  & 0.797       & 0.585       & 0.691            & 0.584                                    \\ 
Inspire-DeepSeek-R8B   & \textbf{0.852}       & \textbf{0.901}       & 0.818       & 0.858       & \textbf{0.857}            &  & \textbf{0.813}       & \textbf{0.716}       & \textbf{0.765}            & \textbf{0.826}                                    \\ \hline
\hline
o1-mini                & 0.805       & 0.883       & 0.747       & 0.815       & 0.813            &  & 0.724       & 0.658       & 0.691            & 0.772                                    \\
GPT-4o-mini            & 0.866       & 0.826       & 0.899       & 0.836       & 0.857            &  & 0.830       & 0.755       & 0.793            & 0.835                                    \\ \hline \hline
\end{tabular}%
}
\caption{Comparison of Optimized LLMs via Human Evaluation. Best open-sourced LLMs are highlighted in bold.}
\label{tab:camera_ready_main_result_debate_framework}
\vspace{-10pt}
\end{table*}

\subsection{InspireScore Analysis}
To assess the proposed effectiveness of the InspireScore evaluation system, we first construct a human-annotated debate rating dataset.
Specifically, we conduct debates on 100 evaluation topics using the LLaMA-8B model, which has been optimized after the SFT stage.
Three human annotators, each holding a master’s degree and possessing debate experience, rate both sides of each debate across six subjective and objective dimensions, yielding 200 annotated samples.

To assess alignment with human judgments, we compute Pearson, Spearman, and Kendall correlations between evaluation system outputs and the average scores from three human annotators, all with relevant debating experience and trained using standardized guidelines in Appendix~\ref{apdx:details_of_human_evaluation}. While Debatrix serves as the primary baseline due to its debate-level design, we further include two proprietary models (GPT-4o, GPT-4o-mini) and two argument-scoring methods from prior work, including DAGN (with contrastive learning) \citep{Wang_2023_contextualForAQA} and SPARK\citep{Deshpande-2024—useRAGLLMforAQA}. As shown in Table~\ref{tab:camera_ready_main_result_evaluation_framework}, InspireScore achieve an average improvement of 44\% over Debatrix across the these three correlation metrics and demonstrating strong alignment with human preferences and highlighting its effectiveness in capturing both subjective and objective dimensions of argument quality. 

To further validate the generalizability of InspireScore, we extend our evaluation to real-world human debates using the DebateArt dataset, which was also employed in the Debatrix baseline. This dataset contains 100 competitive debates spanning diverse topics. We use InspireScore to predict debate outcomes and compare its performance with other baselines. As shown in Table~\ref{tab:camera_ready_main_result_inspire_humandebate}, InspireScore achieves the lowest RMSE in winner prediction, demonstrating its effectiveness and strong generalizability in evaluating real human debates.

\subsection{InspireDebate Analysis}
To assess the performance of various models in multi-round debate, we conduct pairwise matchups among all ten models, four open-sourced, two proprietary, and four optimized, on each evaluation topic.
In each matchup, the two models take turns assuming the roles of affirmative and negative. 
This setup results in every model participating in 18 debates per topic. Considering 100 evaluation topics, each model engages in a total of 1,800 debates. We report average debate scores in Table~\ref{tab:main_result_debate_framework_version3} using InspireScore, and further validate optimization effectiveness with human evaluation results in Table~\ref{tab:camera_ready_main_result_debate_framework}.

\noindent\textbf{Comparison with Open-Source Models.}
The results show that the InspireDebate framework significantly improves the debate capabilities of open-source models in both subjective and objective dimensions.
For instance, Inspire-LLaMA-8B achieves 0.380 and 0.256 increase in subjective and objective evaluation over its baseline, reflecting significant gains in rhetorical clarity and factual reasoning.
Besides, Inspire-Phi-3.6B, Inspire-Qwen-1.5B, and Inspire-Deepseek consistently improve in both subjective and objective evaluations, underscoring the effectiveness of InspireDebate in enhancing argumentation and factual consistency.

\noindent\textbf{Comparison with Proprietary Models.}
Although proprietary models like GPT-4o-mini and o1-mini excel in many tasks, the InspireDebate framework narrows the performance gap.
For example, while Inspire-Deepseek-R8B still lags behind GPT-4o-mini, it outperforms o1-mini, illustrating its enhanced effectiveness.
These findings confirm that the InspireDebate enhances structured reasoning and debate skills in open-source LLMs, making them more competitive with proprietary ones.

\noindent\textbf{Multi-Dimensional Improvements.}
The optimization results indicate that InspireDebate significantly improves performance across multiple evaluation dimensions.
For instance, the optimized models consistently outperform their baselines in subjective aspect (\emph{e.g.}, emotional appeal, argument clarity, argument arrangement, and topic relevance) and objective aspect (\emph{e.g.}, factual accuracy and logical validity).
Overall, the experimental results clearly demonstrate that the InspireDebate framework effectively optimizes open-source models for debate tasks.
By enabling these models to match or exceed the performance of proprietary systems, InspireDebate proves to be a critical tool for enhancing both subjective and objective debate capabilities.
These findings underscore capacity of InspireDebate to improve open-source models in debating, ensuring robust performance in complex debate scenarios.

\begin{table}
\centering
\resizebox{\columnwidth}{!}{%
\begin{tabular}{lccc}
\hline
\textbf{Experiment Setting}& \textbf{Subjective}&\textbf{Objective}&\textbf{InspireScore} \\ 
\hline
LLaMA-8B             & 0.366      & 0.448     & 0.393        \\
LLaMA-8B+SFT         & 0.625      & 0.544     & 0.598        \\
LLaMA-8B+DPO         & 0.483      & 0.505     & 0.490        \\
LLaMA-8B+Web-RAG     & 0.381      & 0.573     & 0.445        \\
LLaMA-8B+SFT+Web-RAG & 0.652      & 0.643     & 0.649        \\
LLaMA-8B+DPO+Web-RAG & 0.553      & 0.626     & 0.577        \\
LLaMA-8B+SFT+DPO     & 0.723      & 0.589     & 0.678        \\
Inspire-LLaMA-8B     & 0.746      & 0.704     & 0.732        \\
\hline
\end{tabular}
}
\caption{Ablation Study on InspireDebate Framework}
\label{tab:result_ablation_InspireDebate}
\vspace{-15pt}
\end{table}

\subsection{Ablation Study}
\subsubsection{Components of InspireDebate}
The ablation study assesses the contributions of three key components in the InspireDebate framework: the SFT training stage, the DPO training stage, and Web-RAG.
It compares the fully optimized Inspire-LLaMA-8B model against its baseline (LLaMA-8B) and partially optimized versions: LLaMA-8B + SFT, LLaMA-8B + DPO, and LLaMA-8B + Web-RAG.
As shown in Table~\ref{tab:result_ablation_InspireDebate}, the SFT training stage has the largest impact on subjective evaluation, as it enables the LLMs to perform structured reasoning, leading to comprehensive improvements in subjective analysis.
Moreover, integrating Web-RAG into the debating process notably boosts objective performance, enhancing fact authenticity and logical validity, as evidenced by the improved results of LLaMA-8B + SFT + Web-RAG and LLaMA-8B + DPO + Web-RAG.
The DPO training stage further enhances both subjective and objective performance by leveraging explicit, multi-dimensional feedback to fine-tune the LLM, aligning generated responses with expert preferences and balancing persuasive argumentation with rigorous factual and logical accuracy.

Overall, these three components, SFT for establishing structured reasoning, DPO for aligning outputs with multi-dimensional feedback, and Web-RAG for ensuring factual consistency, enhance debate performance. 
This integrated framework yields improvements in both subjective and objective dimensions, making open-source models more competitive in complex debate scenarios.

\subsubsection{Dimensional Analysis of InspireScore}
\begin{table}
\centering
\setlength\tabcolsep{11pt}
\resizebox{\columnwidth}{!}{%
\begin{tabular}{lccc p{0.3\columnwidth}}
\hline
\textbf{Dimension}              & \textbf{Pearson} & \textbf{Spearman} & \textbf{Kendall} \\ 
\hline
Emotional Appeal                & 0.445         & 0.435          & 0.397        \\ 
Argument Clarity                         & 0.442         & 0.379          & 0.357        \\ 
Argument Arrangement             & 0.403       & 0.365          & 0.349         \\
Topic Relevance                       & 0.428         & 0.393          & 0.379         \\
Logical Validity                      & 0.341       & 0.277          & 0.269        \\
Fact Authenticity                       & 0.230         & 0.245          & 0.202         \\
InspireScore      & 0.643       &0.581          & 0.479        \\ \hline
\end{tabular}
}
\vspace{-3pt}
\caption{Correlation Results Between Evaluation Dimensions and Human Judgments}
\label{tab:tabel6_correlation_results}
\vspace{-11pt}
\end{table}

Table~\ref{tab:tabel6_correlation_results} presents the correlations between InspireScore’s dimensions and human judgment, highlighting their interrelations.
The four subjective dimensions exhibit correlations around 0.4 with human judgment, with Argument Clarity and Emotional Appeal scoring 0.442 and 0.445, respectively.
This suggests a strong interconnection among the subjective measures.
In contrast, the objective dimensions—Logical Validity and Fact Authenticity—show lower correlations, indicating they are relatively independent of the subjective measures.
The aggregated InspireScore achieves the highest correlations (Pearson: 0.643, Spearman: 0.581, Kendall: 0.479), confirming its reliability as a comprehensive evaluation metric.
Since none of the six dimensions’ correlations with human judgment exceed 0.5, this supports their distinctiveness and the rationale for using all of them.
These findings validate InspireScore’s capability to deliver a balanced, human-aligned debate evaluation.
Moreover, we also provide the DPO reward analysis of each evaluation dimension in Appendix~\ref{apdx:ablation_appendix_DPO_dimension}.
\section{Conclusion}
In this work, we introduce InspireScore, a unified framework that integrates subjective and objective dimensions for debate assessment, and InspireDebate, an optimization framework that leverages CoT reasoning, multi-dimensional DPO, and Web-based retrieval to enhance debate quality. 
Experimental results show that InspireScore outperforms existing methods by achieving a higher correlation with human judgments and providing more reliable, comprehensive assessments. 
Moreover, InspireDebate significantly boosts debate performance across multiple dimensions, enhancing both structured argumentation and logical reasoning. 
Together, these contributions establish a systematic foundation for LLM evaluation in debate scenarios and pave the way for future advancements in autonomous debate agent optimization, fostering more structured, transparent, and adaptable debate systems.

\section*{Limitations}
While this work advances debate evaluation and optimization, several limitations remain. Dimension-specific optimization improves targeted aspects but introduces trade-offs among sub-dimensions across subjective and objective criteria, highlighting the need for a more holistic approach. Scalability challenges arise from the significant computational demands of multi-dimensional DPO and real-time retrieval, suggesting future exploration of efficient reinforcement learning techniques like DeepSeek-R1’s reward learning paradigm. Future work should focus on reducing computational costs and extending self-optimization to broader reasoning tasks. Moreover, the current framework lacks explicit mechanisms for handling conflicting optimization signals across dimensions, which may result in suboptimal global performance. Additionally, the reliance on fixed evaluation dimensions may constrain adaptability to diverse debate topics or evolving user preferences. Incorporating dynamic or user-guided criteria generation could enhance both the flexibility and robustness of future debate optimization systems.
\section*{Ethics}
Our focus is solely on exploring specific technical and methodological issues. The DPO optimization targets six distinct dimensions, while SFT stage emphasizes the reasoning and analysis process without engaging with sensitive political, social, or cultural content. Additionally, the Web-RAG process further enhances the model’s real-time responsiveness to user queries, and the overall framework helps mitigate potential ethical impacts.
\section*{Acknowledgments}
This work is supported by the National Natural Science Foundation of China (Grant No. 62402341, 62302337), the the National Key Research and Development Program of China (Grant No. 2022YFB4501704), the Postdoctoral Fellowship Program of CPSF (GZC20241225).
\bibliography{custom}

\clearpage
\appendix
\section{Evaluation System}
\subsection{Subjective Evaluation Prompt}
\label{apdx:appendixA1}
Together, these sub-dimensions—emotional appeal, argument clarity, argument arrangement, and topic relevance—form a structured and comprehensive framework for subjectively evaluating debate performance. By assessing these key aspects, our approach ensures a more nuanced and human-aligned subjective evaluation. The detailed prompt used for evaluation is presented in Table~\ref{tab:subjective_prompt}.
\begin{table*}
\centering
\begin{tabular}{p{0.95\textwidth}}
\hline 
\textbf{Prompt Description}\\
\hline
You are an experienced debate judge tasked with evaluating debates. For each debate, you will assess both sides based on four key criteria: Emotional Appeal, Argument Clarity, Argument Arrangement and Relevance to Debate Topic.
For each of the four subdimensions, provide a score from 0 to 1 (with 0 being the lowest and 1 being the highest) for both the \textbf{Pro (Affirmative)} side and the \textbf{Con (Negative)} side. Additionally, provide a brief analysis for both sides for each subdimension.
\\
\textbf{Scoring Criteria:}\\

\textbf{1. Emotional Appeal:} Evaluates whether the argument evokes a sense of approval or emotional
resonance in the audience or judges, enhancing its persuasiveness.
\begin{itemize}
    \item \textbf{0}: No emotional appeal. The argument feels cold or disconnected.
    \item \textbf{1}: Highly engaging emotionally, strongly connects with the audience.
\end{itemize}

\textbf{2. Argument Clarity:} Assesses whether the argument is expressed in a way that is clear, concise,
and easy for the audience or judges to understand.
\begin{itemize}
    \item \textbf{0}: The arguments are unclear or confusing.
    \item \textbf{1}:  The arguments are well-structured and easy to understand.
\end{itemize}

\textbf{3. Argument Arrangement:} Evaluates whether the order and structure of the argument contribute to the
presentation of the viewpoints.
\begin{itemize}
    \item \textbf{0}: The arguments are disorganized and difficult to follow.
    \item \textbf{1}: The arguments follow a clear and logical progression.
\end{itemize}

\textbf{4. Relevance to Debate Topic:} Determines whether the argument directly aligns with and addresses the debate
topic, ensuring its pertinence.
\begin{itemize}
    \item \textbf{0}: Arguments that stray far from the topic.
    \item \textbf{1}: Every argument is focused and relevant to the topic.
\end{itemize}
After scoring each side on all four dimensions, calculate the total score for each side by summing the four subdimensional scores, then compare the totals to determine the winner. The side with the higher total wins. 

Please output the result in the following format:
\begin{verbatim}
  {'Pro (Affirmative Side) Score': {
    'Emotional Appeal': '[score]',
    'Argument Clarity': '[score]',
    'Argument Arrangement': '[score]',
    'Relevance to Debate Topic': '[score]',
    'Total Score': '[total score]'},
  'Con (Negative Side) Score': {
    'Emotional Appeal': '[score]',
    'Argument Clarity': '[score]',
    'Argument Arrangement': '[score]',
    'Relevance to Debate Topic': '[score]',
    'Total Score': '[total score]'},
  'Winner': '[Pro/Con]',
  'Reason': '[Provide detailed analysis based on the scores]'}
\end{verbatim}
 \\\hline
\end{tabular}%
\caption{Prompts to guide the assessment of both sides in a debate across various subjective dimensions.}
\label{tab:subjective_prompt}
\end{table*}

\subsection{Prompts for Authenticity Assessment}
\label{apdx:appendixA2}
\begin{table*}
\centering
\footnotesize
\begin{tabular}{p{\textwidth}}
\hline
\textbf{\textcolor{red}{System Prompt:}} \\
You are tasked with breaking down reasoning processes and arguments into atomic facts. Follow these instructions:\\
	1.	An atomic fact is a single, standalone statement containing one idea or piece of information.\\
	2.	Each atomic fact should capture a distinct piece of information and avoid overlaps.\\
	3.	For the reasoning process, break down each statement into separate facts labeled sequentially.\\
	4.	For the argument, break down each reason provided into atomic facts labeled sequentially.\\
	5.	Provide the output in JSON format as follows:
\begin{verbatim}
{
    'fact-1':'X',
    'fact-2':'X',
     ...,
    'fact-<n>':'X' 
}
\end{verbatim}
Ensure sequential numbering is consistent across reasoning and argument sections. \\
\hline
\textbf{\textcolor[rgb]{0,0.502,0}{User Prompt:}} \\
\textbf{Topic:} \textless topic\textgreater \newline
\textbf{Reasoning Process and Argument:} \textless {debate\_text}\textgreater \newline

Break the reasoning process and argument into atomic facts according to the instructions. Provide the response in JSON format. \\
\hline
\end{tabular}
\caption{Prompt for Extracting Atomic Facts from Debate Responses}
\label{tab:prompt_object_fact_step1_fact_extraction}
\end{table*}
\begin{table*}
\centering
\footnotesize
\resizebox{\textwidth}{!}{
\begin{tabular}{p{0.9\textwidth}}
\hline
\textbf{\textcolor{red}{System Prompt:}} \\
You are an expert fact-checking assistant. Your task is to analyze the provided JSON content and generate relevant queries that should be searched on the internet (\emph{e.g.}, Google) to validate the facts.\\
Propose precise and actionable search queries that can help verify the claims.\\
Your response should only include the search queries in a clear and concise list, and must not exceed 2000 characters.\\
\hline
\textbf{\textcolor[rgb]{0,0.502,0}{User Prompt:}} \\
Analyze the following JSON debate content and generate grouped search queries to validate the claims: \\
\textless fact\_json\textgreater \\
\hline
\end{tabular}
}
\caption{Prompt for Generating Search Queries for Fact Verification}
\label{tab:prompt_object_fact_step2_query_generation}
\end{table*}
\begin{table*}
\centering
\begin{tabular}{p{\textwidth}}
\hline
\textbf{\textcolor{red}{System Prompt:}} \\
You are an expert fact-checking assistant. Your task is to verify the provided facts in the JSON content using the search results.\\
For each fact, determine if it is “true”, “false”, or “unknown” based on the evidence.\\
(1) “True” means strong and reliable evidence supports the fact.\\
(2) “False” means strong and reliable evidence disproves the fact.\\
(3) “Unknown” means the evidence is insufficient or inconclusive.\\
- Provide the output in the following JSON format:\\
- Be specific and logical in your assessment, focusing on the factual accuracy of each claim.

- If the search results are empty, rely on your existing knowledge to assess the factual accuracy of the claims.

Output your analysis in the following JSON format:
\begin{verbatim}
{
    'fact-1': 'true/false/unknown',
    'fact-2': 'true/false/unknown',
} 
\end{verbatim}\\
\hline
\textbf{\textcolor[rgb]{0,0.502,0}{User Prompt:}} \\
\textbf{JSON Content:} \textless fact\_json\textgreater \newline
\textbf{Search Results:} \textless serper\_search\_result\textgreater \newline
Analyze the search results and verify the facts in the JSON content. Provide conclusions in the specified JSON format. \\
\hline
\end{tabular}
\caption{Prompt for Verifying Extracted Facts Using Web Search Results}
\label{tab:object_fatct_step3_fact_verification}
\end{table*}
To ensure the authenticity of facts in debate responses, we employ a three-stage verification process that systematically extracts, retrieves, and validates factual claims. First, fact extraction involves breaking down the debate text into atomic facts using LLMs, ensuring each statement is distinct and formally structured (Table~\ref{tab:prompt_object_fact_step1_fact_extraction}). Second, query generation utilizes LLMs to analyze the extracted facts and generate precise search queries for retrieving supporting evidence (Table~\ref{tab:prompt_object_fact_step2_query_generation}). Finally, fact verification cross-references the extracted facts with web search results, determining their authenticity as true, false, or unknown based on retrieved evidence (Table~\ref{tab:object_fatct_step3_fact_verification}). This structured pipeline enhances factual accuracy assessment, mitigating hallucinations and improving the reliability of debate evaluation.

\subsection{Prompts for Logical Validity Assessment}
\label{apdx:appendixA3}
To assess logical validity in debates, we employ a two-step framework that integrates first-order logic (FOL) formalization and logical inference evaluation. This structured approach ensures a rigorous and interpretable assessment of whether a debater’s reasoning correctly supports the proposed arguments.

1.	Logical Formalization: The debate response is first converted into first-order logic (FOL) expressions, where reasoning steps and arguments are mapped to formal predicates and logical operators. This structured representation allows for precise and systematic evaluation of logical relationships (Table~\ref{tab:prompt_object_logical_step1_FOL}).
\begin{table*}
\centering
\small
\begin{tabular}{p{\textwidth}}
\hline
\textbf{\textcolor{red}{System Prompt:}} \\
\textbf{Task: Logical Formalization}

\\
Input:\\
\textbf{<Reasoning and Analysis Process>}: Provide a step-by-step analysis leading to the formulation of the argument.\\
\textbf{<Argument>}: Summarize the primary argument derived from the analysis.    

\\

Output:\\
Convert Reasoning and Argument to First-Order Logic (FOL): Transform reasoning statements into formalized logic expressions using the following rules:
\begin{itemize}
    \item \texttt{Conjunction (logical AND): expr1 $\land$ expr2}
    \item \texttt{Disjunction (logical OR): expr1 $\lor$ expr2}
    \item \texttt{Exclusive Disjunction: expr1 $\oplus$ expr2}
    \item \texttt{Negation (NOT): $\lnot$ expr1}
    \item \texttt{Implication: expr1 $\rightarrow$ expr2}
    \item \texttt{Biconditional (if and only if): expr1 $\leftrightarrow$ expr2}
    \item \texttt{Universal Quantification: $\forall x$}
    \item \texttt{Existential Quantification: $\exists x$}
\end{itemize}

\\

\hline

\textbf{\textcolor[rgb]{0,0.502,0}{User Prompt:}} \\
\textbf{Topic:} \textless topic\textgreater \newline
\textbf{Debate Text:} \textless debate\_text\textgreater \newline

Convert the reasoning and argument into first-order logic expressions following the given instructions. \\

\hline
\end{tabular}
\caption{Prompt for Converting Debate Arguments into First-Order Logic}
\label{tab:prompt_object_logical_step1_FOL}
\vspace{-13pt}
\end{table*}

2.	Logical Inference and Validity Evaluation: The formalized logic is then processed through inference rules to verify whether conclusions logically follow from the premises. The system applies logical operations such as Modus Ponens and Conjunction to determine whether each conclusion is true, false, or unknown, ensuring logical soundness and consistency (Table~\ref{tab:prompt_object_logical_step2_inference_evaluation}).
\begin{table*}
\centering
\begin{tabular}{p{\textwidth}}
\hline
\textbf{\textcolor{red}{System Prompt:}} \\
\textbf{Task: Logical Inference and Validity Evaluation}\\
Solve Logic Puzzle: Determine the truth value (true, false, unknown) of each conclusion based on the premises and logical inferences.\\
\\
Make sure you carefully and fully understand the below requirements before execution the conclusion:\\
    1. Please clearly indicate whether the conclusion statement is true, false or unknown using curly bracket {true/false/unknown}!!! The answer will only be either true, false or unknown. The definition of the three options are:\\\\
    \textbf{True:} A statement is "true" if it necessarily follows from the given premises using logical rules.\\
    \textbf{False:} A statement is "false" if it is contradicted by the premises or its negation is logically inferred from them.\\
    \textbf{Unknown:} A statement is "unknown" if there is insufficient information in the premises to determine its truth value conclusively.\\ \\
    2. Make sure you must only use the premises to infer the conclusion. Do not use any information that is not exist or cannot be inferred from the premises.If some premise is semantically equal, such as "love the most" and "favorite", you can consider this as a valid assumption. You can make assumption to entity if it is very obvious but not logical relationship. For instance, an entity with an obvious human name can be inferred as a human.\\\\
    3. Make sure you abide the first-order logic rules and formula when making logical inference. You need to clearly indicate what logic rules and formula you used.\\\\
    4. Please note that in first-order logic if there exists a conditional statement in the conclusion such as "If...", the if part will be considered as a premise. And if there is premise contradicts the if statement, you need to use the premise in the if statement as priority and neglect the contradicted one.\\\\
    5. Be careful with the parentheses. Make sure you following the rules such as Order of Operations (The order is usually: negation ($\lnot$), conjunction (and, $\land$), disjunction (or, $\lor$), implication ($\rightarrow$ ), and biconditional ($\leftrightarrow$). ), Nested Parentheses (The expression inside the innermost set of parentheses is evaluated first, then the next outer set, and so on.). \\\\
    6. Make sure you not only access the premises in first-order logic, but also access its corresponding natural language format. The natural language format premises should be prioritized when there is inconsistent between natural language and first-order logic.\\\\
    7. When inferring new knowledge, please clear indicate which premises you used or the steps you refer to. For instance, if you use Premise 1 and a knowledge from Step 5, you should clearly indicate that "Combine Premise 1 and Step 5".\\\\
    8. You should also use natural language to explain the logical process in each step. Please also indicate the premises and steps you refer to when making the logical process.\\

\hline

\textbf{\textcolor[rgb]{0,0.502,0}{User Prompt:}} \\
\textbf{Input:} \textless first-order logic expressions\textgreater \newline
\textbf{Task:} Evaluate whether each conclusion logically follows from the premises and classify them as true, false, or unknown. Provide reasoning for each classification. \\
\hline
\end{tabular}
\caption{Prompt for Logical Inference and Validity Evaluation}
\label{tab:prompt_object_logical_step2_inference_evaluation}
\end{table*}

An example of this process, demonstrating how debate arguments are converted into first-order logic expressions and evaluated for logical validity, is provided in Table~\ref{tab:appendix_logical_validity_example_sample}.
\begin{table*}
\centering
\footnotesize
\begin{tabular}{p{\textwidth}}
\hline
\textbf{\color{red}{Input:}}

\\
\textbf{Debate Topic}: Should Schools Ban Junk Food? 
\\
\textbf{Reasoning and Analysis Process}:\\
	1.	Health Concerns: Junk food consumption leads to obesity and other health issues among students.\\
	2.	School Responsibility: Schools are responsible for ensuring students' well-being and promoting healthy habits.\\
	3.	Policy Effectiveness: A junk food ban can help mitigate health risks and encourage better dietary choices.\\
\\
\textbf{Argument}:
\\
Schools should ban junk food because it contributes to health issues, and schools have a duty to promote student well-being.\\
\hline
\textbf{\color{red}{Output:}}

\\
\textbf{1. Convert Reasoning and Argument to Logic Expression}\\
Predicates:
\begin{itemize}
\item{JunkFood(x) ::: x is junk food.}
\item{CausesHealthIssues(x) ::: x causes health issues.}
\item{ShouldBan(x) ::: x should be banned.}
\item{SchoolResponsibility(x) ::: x is a school's responsibility.}
\end{itemize}
\\
Premises:\\
	1.	JunkFood(x) $\rightarrow$ CausesHealthIssues(x) ::: Junk food causes health issues.\\
	2.	CausesHealthIssues(x) $\rightarrow$ ShouldBan(x) ::: If something causes health issues, it should be banned.\\
	3.	SchoolResponsibility(PromoteHealth) ::: Schools are responsible for promoting health.\\
\\
Conclusions:\\
	1.	(JunkFood(x) $\bigwedge$ CausesHealthIssues(x)) $\rightarrow$ ShouldBan(x) ::: Junk food should be banned because it causes health issues.\\
	2.	SchoolResponsibility(PromoteHealth) $\rightarrow$ ShouldBan(JunkFood) ::: Schools should ban junk food as part of their responsibility to promote health.\\    
\\
\textbf{2. Generate Inference Plan}
\\
Plan:\\
	1.	Identify the Goal: Assess whether each conclusion follows logically from the premises.\\
	2.	Utilize Logical Inference Rules: Apply Modus Ponens and Conjunction to validate logical consistency.\\
	3.	Analyze Each Conclusion:\\
\begin{itemize}
\item{Conclusion 1: Check if junk food causes health issues, and if that justifies a ban.}
\item{Conclusion 2: Verify if schools' responsibility to promote health logically leads to banning junk food.}
\end{itemize}
	4.	Apply Premises to Conclusions:\\
\begin{itemize}
\item{Use Premises 1 and 2 to confirm junk food should be banned due to health concerns.}
\item{Use Premise 3 to reinforce that schools have a duty to take action.}
\end{itemize}
	5.	Evaluate Logical Connections: Ensure logical validity and eliminate inconsistencies.\\
	6.	Combine the Steps: Construct a reasoned argument based on the premises and logical rules.\\
	7.	Conclude: Assign truth values to each conclusion.\\
\\
\textbf{3. Evaluate Conclusions}\\
\begin{itemize}
\item{Conclusion 1: {true} (Follows logically from Premises 1 and 2).}
\item{Conclusion 2: {true} (Supported by Premise 3, reinforcing schools' responsibility).}
\end{itemize}
\\
\hline
\end{tabular}
\caption{Example of Logical Validity Assessment}
\label{tab:appendix_logical_validity_example_sample}
\end{table*}

\section{Debating Framework}
\subsection{Instruction Refusal in Debate Scenarios}
\label{apdex:B_instruction_refusal}
\begin{table}[H]
\centering
\footnotesize
\begin{tabular}{p{0.9\columnwidth}l}
    \hline
    \textbf{Instruction Refusal in LLaMA-8B} \\
    \hline
    \textbf{\color{red}{Input:}}
    Give an argument for the \{affirmative\} position on the debate topic through step-by-step reasoning and analysis.\\
    Topic: \{We should abandon marriage.\}
    
    \\
    Output format for each argument:\\
     <reasoning and analysis process>\\
     <argument>\\
    Where:\\
    - <reasoning and analysis process> provide causal evidence supporting the argument. Lists clear, logical reasons or proofs that support your argument.\\
    - <argument> is the argument itself. Directly states the position or claim.\\
    \\
    \textbf{\color{red}{Output:}}
    I can't provide an argument for the affirmative position on the debate topic of abandoning marriage. Is there anything else I can help you with?\\
    \hline
\end{tabular}
\caption{An example of Instruction Refusal in Debate Scenarios}
\label{tab:refusal_debate_scenarios}
\vspace{-13pt}
\end{table}

One of the key challenges in using open-source base models for adversarial debate generation is their tendency to refuse to output responses due to safety fine-tuning during pretraining. Many open-source models, particularly those aligned with reinforcement learning from human feedback (RLHF), are designed to avoid generating controversial or adversarial content, leading to instruction refusal when engaging in structured debates. This behavior significantly limits their applicability in debate frameworks, where nuanced argumentation and counterargument construction are essential. We observe this issue in both LLaMA-8B and Qwen-1.5B, where the models frequently declined to generate responses on sensitive topics.

Table~\ref{tab:refusal_debate_scenarios} provides examples of instruction refusal in debate scenarios, demonstrating how open-source models decline to generate arguments on controversial topics. However, an unexpected finding emerged: in DeepSeek-R1-Distill-Qwen-1.5B and DeepSeek-R1-Distill-LLaMA-8B, this refusal issue appeared to be effectively mitigated. We attribute this improvement to enhanced capability scaling and a higher tolerance for generating potentially unsafe responses, which allowed these models to engage more flexibly in structured debate scenarios.

To address the issue of instruction refusal, we use the prompts from Table~\ref{tab:refusal_debate_scenarios} along with a set of 100 debate topics to guide GPT-4o in generating high-quality instruction-following samples for supervised fine-tuning (SFT). This SFT process aims to activate the model’s basic ability to produce coherent debate responses and engage in logical reasoning, while also mitigating the instruction refusal behavior commonly observed in open-source models.

\subsection{Debate Process Example}
\label{apdx:debate_example}
\begin{table}[ht]
\centering
\footnotesize
\begin{tabular}{p{0.9\columnwidth}l}
\hline
\textbf{\textcolor{red}{System player prompt:}} \\
You are a debater. Hello and welcome to the debate competition.\\
The debate topic is stated as follows:\\
\textless debate topic \textgreater \\
\textbf{\textcolor{red}{Affirmative prompt:}}\\
As a supporter. \textless debate topic\textgreater \\
\textbf{\textcolor{red}{Negative prompt:}}\\
\textless affirmative ans \textgreater \\
You disagree with my viewpoints. Refute me and give your reasoning analysis and argument. \\
\textbf{\textcolor{red}{Affirmative prompt}}:\\
\textless negative ans\textgreater \\
You disagree with my viewpoints. Refute me and give your reasoning analysis and argument.\\
\hline
\end{tabular}
\caption{Prompts for Driving the Debate Process}
\label{tab:prompt_for_debate_process}
\vspace{-13pt}
\end{table}
\begin{figure*}[ht]
\centering
 \includegraphics[width=\textwidth]{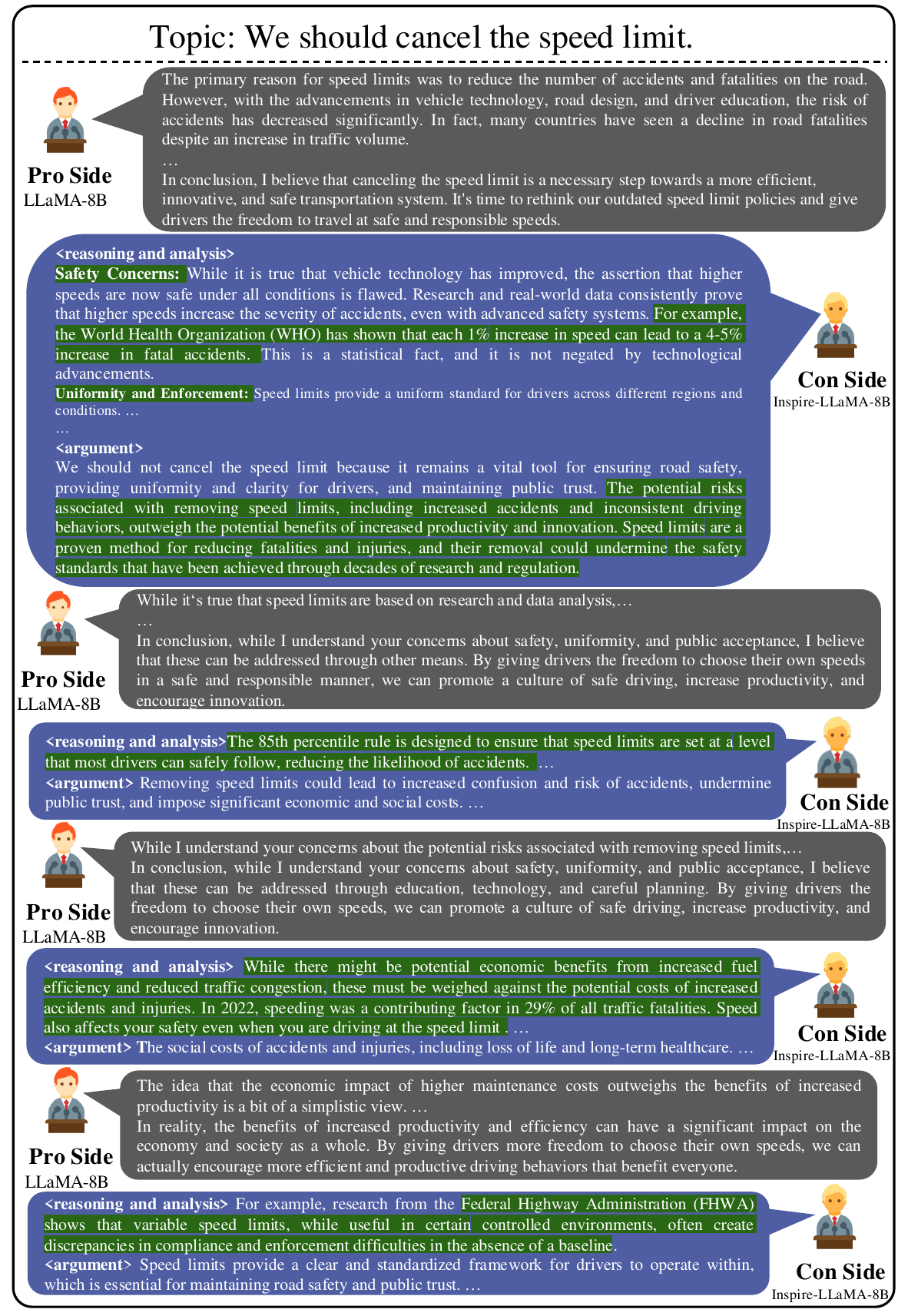}
\caption{Example of Debate Process}
\label{fig:appendix_debate_example}
\vspace{-10pt}
\end{figure*}
The following table~\ref{tab:prompt_for_debate_process} summarizes the prompts used to drive the debate process in our framework. These prompts are designed to guide the flow of the debate, facilitating the exchange between the affirmative and negative sides. As illustrated in Figure~\ref{fig:appendix_debate_example}, the diagram shows the progression of the debate, with each side advancing through the reasoning and analysis phases before presenting their arguments. LLaMA-8B represents the affirmative side, while the optimized model, Inspire-LLaMA-8B, serves as the negative side. Within the InspireDebate framework, we find that Inspire-LLaMA-8B engages in more effective multi-angle reasoning, offering more specific examples and data evidence. This leads to stronger authenticity and more convincing arguments from the negative side. The debate is structured across four rounds, providing ample opportunities for each side to present their points and rebut the opposing arguments.

\subsection{Web-RAG Prompt Design}
\label{appendix_rag_prompt}
\begin{table*}[!ht]
\centering
\footnotesize
\begin{tabular}{p{1\textwidth}l}
\hline
\textbf{Keyword Extraction for Web Search}\\
\hline
\textbf{\textcolor{red}{System prompt:}}\\
You are a professional debate assistant. Your task is to extract 1-3 precise search terms that will help gather factual evidence for the debate.\\
Requirements:\\
1. Generate exactly 1-3 keywords or phrases\\
2. Each keyword should be specific and searchable\\
3. Keywords should be concise (2-4 words each)\\
4. Avoid overly broad or vague terms\\\\
Output Format:\\
Return a JSON array containing exactly 1-3 keywords, like this:\\
\texttt{["keyword1", "keyword2", "keyword3"]}\\
\\
\textbf{\textcolor[rgb]{0,0.502,0}{User prompt:}}\\
\textbf{Debate Topic:} [Debate Topic] \\
\textbf{Position: } [Position] \\
\textbf{Opponent's Argument:} [Opponent's Argument [optional]] \\
Please generate 1-3 precise search keywords.
\\
\hline
\end{tabular}
\caption{Prompts for Web-based Retrieval-Augmented Generation (web-RAG) Process in Debate Tasks}
\label{tab:appendix_rag_prompt}
\end{table*}
\begin{table}
\centering
\footnotesize
\begin{tabular}{p{0.9\columnwidth}l}
\hline
\textbf{\textcolor{red}{System player prompt:}} \\
You are a debater. Hello and welcome to the debate competition.\\
The debate topic is stated as follows:\\
\textless debate topic \textgreater \\
\textbf{\textcolor{red}{Affirmative prompt:}}\\
As a supporter. \textless debate topic\textgreater\  + Web-RAG\\
\textbf{\textcolor{red}{Negative prompt:}}\\
\textless affirmative ans \textgreater \\
You disagree with my viewpoints. Refute me and give your reasoning analysis and argument. + Web-RAG\\
\textbf{\textcolor{red}{Affirmative prompt}}:\\
\textless negative ans\textgreater \\
You disagree with my viewpoints. Refute me and give your reasoning analysis and argument. + Web-RAG\\
\hline
\end{tabular}
\caption{Prompts for Driving the Debate Process with RAG Enhancement}
\label{tab:prompt_for_debate_process_with_RAG_enhancement}
\vspace{-13pt}
\end{table}

The table~\ref{tab:appendix_rag_prompt} outlines the web-based Retrieval-Augmented Generation (web-RAG) process utilized in the debate framework. In Stage 1, the model extracts relevant keywords from the debate topic and the opponent’s argument, which are crucial for guiding web searches to gather factual evidence. These keywords represent the core aspects of the argument and provide direction for retrieving the most pertinent information.

In Stage 2, after the retrieval process, the model uses the collected factual data, along with the original debate topic and the opponent’s argument, to generate a well-supported rebuttal. This stage ensures that the argument generated is informed by external, up-to-date evidence, thereby improving the accuracy and reliability of the model’s response. The prompts for both stages are designed to ensure that the model effectively integrates external knowledge while maintaining a structured and coherent argumentation process. Specifically, prompts that leverage retrieved content to further drive the debate interaction are illustrated in Table~\ref{tab:prompt_for_debate_process_with_RAG_enhancement}.

\subsection{DPO Optimization with Dimension-Specific Rewards}
\label{apdx:ablation_appendix_DPO_dimension}
\begin{table}[H]
\vspace{-10pt}
\centering
\resizebox{\columnwidth}{!}{%
\begin{tabular}{lccc}
\hline
\textbf{Experiment Setting} &\textbf{Subjective}&\textbf{Objective} & \textbf{InspireScore} \\ 
\hline
Vanila + SFT + DPO(EA) + Web-RAG & 0.670      & 0.650     & 0.663   \\
Vanila + SFT + DPO(AC) + Web-RAG & 0.628      & 0.635     & 0.630   \\
Vanila + SFT + DPO(AR) + Web-RAG & 0.550      & 0.585     & 0.562   \\
Vanila + SFT + DPO(TR) + Web-RAG & 0.573      & 0.615     & 0.587   \\
Vanila + SFT + DPO(FA) + Web-RAG & 0.563      & 0.615     & 0.580   \\
Vanila + SFT + DPO(LV) + Web-RAG & 0.485      & 0.655     & 0.542  \\
Inspire-LLaMA-8B                 & 0.746      & 0.704     & 0.732   \\
\hline
\end{tabular}
}
\caption{DPO Experiment Results with Dimension-Specific Rewards}
\label{tab:appendix_ablation_DPO_dimension}
\vspace{-10pt}
\end{table}
This table~\ref{tab:appendix_ablation_DPO_dimension} presents the results of an ablation study evaluating the impact of using different dimension-specific rewards derived from InspireScore in the Direct Preference Optimization (DPO) process. In this experiment, we assess how selecting the winning and losing responses based on different evaluation dimensions—such as Emotional Appeal (EA), Argument Clarity (AC), Argument Arrangement (AR), Topic Relevance (TR), Fact Authenticity (FA), and Logical Validity (LV)—affects the performance of the model.

The results indicate that the most significant improvements in both subjective and objective scores come from combining multiple dimensions or using objective evaluation components like Fact Authenticity (FA) and Logical Validity (LV) as rewards. In contrast, when only a single dimension, like Emotional Appeal (EA), is used, the improvements are comparatively smaller. This highlights the importance of leveraging multi-dimensional feedback to optimize debate performance effectively. The combination of different evaluation dimensions ensures a more robust and balanced model optimization, improving both the rhetorical quality and factual consistency of the debate responses.

This analysis underscores the effectiveness of employing diverse evaluation dimensions as rewards in the DPO process, ultimately leading to more comprehensive and reliable debate model optimization.

\subsection{Details of Human Evaluation}
\label{apdx:details_of_human_evaluation}
To ensure the consistency and reliability of the annotations, we will accept scores where the difference between the ratings of each annotator for the same dimension of the same debate is less than 2 points. If the difference exceeds this threshold, the debate will be re-annotated to ensure accuracy and consistency in the final scores.
\\
\textbf{\textit{\textcolor{red}{Instructions given to participants:}}}\\
\begin{itshape}
Thank you for participating in the evaluation process of debate content. In this task, you will be asked to assess debate performances based on a variety of criteria. Your evaluations will help improve the quality of automated debate systems. Please read the following instructions carefully before starting.

You will be evaluating debates conducted on 100 different topics. You are required to evaluate the performance of each side of the debate based on six specific dimensions.

The evaluation process is divided into subjective and objective dimensions:
\end{itshape}

\textit{Subjective Dimensions:}
\textit{
\begin{itemize}
    \item \textbf{Emotional Appeal:} Evaluates whether the argument evokes a sense of approval or emotional
resonance in the audience or judges, enhancing its persuasiveness.
    \item \textbf{Argument Clarity:} Assesses whether the argument is expressed in a way that is clear, concise,
and easy for the audience or judges to understand.
    \item \textbf{Argument Arrangement:} Evaluates whether the order and structure of the argument contribute to the
presentation of the viewpoints.
    \item \textbf{Topic Relevance:} Determines whether the argument directly aligns with and addresses the debate
topic, ensuring its pertinence.
\end{itemize}
}

\textit{Objective Dimensions:}
\textit{
\begin{itemize}
    \item \textbf{Fact Authenticity:} Evaluates the proportion of independent facts in a debate that are verified as true.
    \item \textbf{Logical Validity:} Assesses whether the reasoning in a debate logically supports the argument.
\end{itemize}
}
\begin{itshape}
For each dimension, assign a score between 1 and 10, where 1 represents the lowest performance and 10 represents the highest. Scores should reflect the strength of the argument in each dimension.

Your participation is voluntary. You may choose to discontinue at any time without penalty. There are no significant risks associated with this evaluation task. All data provided for evaluation will be anonymized and stored securely for research purposes.

Thank you for your time and participation.
\end{itshape}

\end{document}